\documentclass{article}
\usepackage{ijcai26}

\usepackage{times}
\usepackage{soul}
\usepackage{url}
\usepackage[hidelinks]{hyperref}
\usepackage[utf8]{inputenc}
\usepackage[small]{caption}
\usepackage{graphicx}
\usepackage{amsmath}
\usepackage{amsthm}
\usepackage{booktabs}
\usepackage{algorithm}
\usepackage{algorithmic}
\usepackage[switch]{lineno}
\usepackage{amsmath,latexsym,amssymb,amsthm,array,amsfonts,algorithm,booktabs,graphicx,subfigure,multirow,cuted,stfloats}
\usepackage{mathrsfs}
\usepackage{amsfonts,amssymb}
\usepackage{threeparttable}
\usepackage{fancyvrb}
\usepackage{tcolorbox}
\usepackage{listings}
\usepackage{fancyhdr}
\usepackage{lipsum} 
\usepackage{wrapfig}
\usepackage{hyperref}
\usepackage{url}
\usepackage{multirow}

\title{LacaDM: A Latent Causal Diffusion Model for Multiobjective Reinforcement Learning}

\author{
Xueming Yan$^1$,
Bo Yin$^{2}$,
Yaochu Jin$^{2*}$
\affiliations
$^1$School of Information Science and Technology, Guangdong University of Foreign Studies\\
$^2$School of Engineering, Westlake University\\
\emails
yanxm@gdufs.edu.cn; yb0927@outlook.com;
jinyaochu@westlake.edu.cn\\
}

\begin{document}

\maketitle

\begin{abstract}
Multiobjective reinforcement learning (MORL) poses significant challenges due to the inherent conflicts between objectives and the difficulty of adapting to dynamic environments. Traditional methods often struggle to generalize effectively, particularly in large and complex state-action spaces. To address these limitations, we introduce the Latent Causal Diffusion Model (LacaDM), a novel approach designed to enhance the adaptability of MORL in discrete and continuous environments. Unlike existing methods that primarily address conflicts between objectives, LacaDM learns latent temporal causal relationships between environmental states and policies, enabling efficient knowledge transfer across diverse MORL scenarios. By embedding these causal structures within a diffusion model-based framework, LacaDM achieves a balance between conflicting objectives while maintaining strong generalization capabilities in previously unseen environments. Empirical evaluations on various tasks from the MOGymnasium framework demonstrate that LacaDM consistently outperforms the state-of-art baselines in terms of hypervolume, sparsity, and expected utility maximization, showcasing its effectiveness in complex multiobjective tasks.
\end{abstract}

\section{Introduction}
Multiobjective reinforcement learning (MORL) \cite{felten2023toolkit,lu2023multi} has become a prominent research area due to its ability to address real-world problems where multiple, often conflicting objectives must be optimized simultaneously. In contrast to traditional reinforcement learning, where a single objective is maximized, MORL requires to balance and optimize multiple objectives, often in dynamic and uncertain environments \cite{gu2025safe}. This adds significant complexity to the learning process, especially when the objectives conflict with each other or change over time. Moreover, large, high-dimensional state-action spaces in many MORL tasks pose scalability challenges to existing algorithms, making it difficult for MORL to generalize across different tasks \cite{felten2024multi}.

Traditional approaches to MORL have primarily focused on scalarization methods \cite{van2013scalarized,zheng2023generalized}, Pareto-based methods \cite{cai2023distributional,liu2025pareto,van2014multi,zheng2022survey}. While these methods show impressive success in solving multiobjective problems, they often struggle to generalize, particularly in complex environments where the relationships between objectives are nonlinear and the state-action space are subject to changes.
Scalarization methods \cite{agarwal2022multi,gu2024large} combine objectives into a weighted sum or other forms of aggregation, but the choice of weights often requires prior knowledge of the relative importance of each objective, which may not be available in real-world scenarios \cite{xia2021combined}. Pareto-based methods \cite{tian2022local}, which rank solutions based on their dominance relationships, often require to maintain a set of nondominated solutions, which can be computationally expensive, especially in high-dimensional spaces. Furthermore, Pareto-based approaches struggle when objectives conflict in complex ways or when the set of nondominated solutions becomes large and difficult to manage \cite{lin2022pareto}.
Additionally, both scalarization and Pareto-based methods \cite{zhang2024evolutionary} typically assume that the objectives remain relatively stationary, which is not always the case in dynamic or uncertain environments. Therefore, these methods often struggle with generalization, particularly in complex environments where the relationships between the objectives are nonlinear, and the state-action space is highly time-varying.

Recently, diffusion models \cite{ho2020denoising,sanokowski2024diffusion,song2020denoising} have demonstrated attractive generalization abilities and been applied to various types of optimization problems. For example, EmoDM \cite{yan2024emodm} adopts a diffusion model to solve multi-objective optimization problems (MOPs) that considers the evolutionary search process as a reverse diffusion process. By pretraining on previously solved MOPs, EmoDM can generate a set of nondominated solutions for a new MOP through reverse diffusion, without the need to perform additional evolutionary search.
A generative diffusion model, called GDMTD3 \cite{zhang2024multi}, was proposed to solve the aerial collaborative secure communication optimization problem in multiobjective reinforcement learning.
While these diffusion models have advanced the field of multiobjective optimization \cite{li2024expensive,zhang2024multi}, they still require substantial training on past optimization tasks to generalize effectively. This reliance on extensive pretraining limits the model's adaptability to new, unseen environments, hindering its scalability and applicability to complex, real-world MORL problems \cite{zeng2024survey}.
Furthermore, most of these diffusion models fail to account for the temporal and latent dependencies that arise from the interaction between the agent’s actions and the evolving environment, which are critical for robust decision-making in dynamic settings and also significantly impact generalization performance \cite{zhang2024multi}. 

Fortunately, we find that the interactive mechanism between the agent and the environment can be characterized by causality. Causal inference is a useful tool for modeling the generative mechanism behind the agent's actions. Inspired by this, we propose a Latent Causal Diffusion Model (LacaDM) for MORL, which incorporates causal relationships between objectives and the environment directly into its latent space, enabling more structured and data-efficient policy generation.
In the forward diffusion process, LacaDM learns to approximate the reserved search process of a deep reinforcement learning algorithm, such as Pareto Conditioned Networks (PCN) \cite{reymond2022pareto}, starting from the optimal policy and gradually evolving toward a random initial policy. In the reverse diffusion process, noise is progressively removed from the random policy, incrementally refining it into an optimal policy approximation. 
The main contributions of this work are as follows:
\begin{itemize}
    \item We propose a latent causal diffusion model to solve MORL problems. The key idea is to integrate causal representation learning  into the diffusion process to enhance generalization across a wide range of MORL tasks. LacaDM can be used as a general diffusion model applicable to both continuous and discrete MORL environments. 
    \item We optimize the policy by learning the latent causal dynamics during the reverse diffusion process, thereby improving LacaDM’s ability to adapt to dynamic environments. This strategy enables LacaDM to continuously learn and adapt to new tasks without the requirement for exhaustive pretraining on previous problems.  
    \item Extensive experiments conducted across a variety of environments from MOGymnasium demonstrate the superiority of LacaDM over existing baseline methods in terms of hypervolume, sparsity and expected utility maximization across a wide range of MORL tasks.
\end{itemize}

\section{Preliminaries}

\subsection{Diffusion Probabilistic Models}
Diffusion models \cite{yang2023diffusion} are generative frameworks that refine noise-corrupted samples to generate high-quality data. They consist of two phases: the forward process, which adds noise, and the reverse process, which removes it to recover the original sample. These models \cite{nie2025erasing,guo2024audio,ye2022first} are classified into two types: continuous diffusion models, for continuous data (e.g., images, audio), and discrete diffusion models, for discrete data (e.g., binary, categorical). While both follow similar principles, they differ in how noise is added and removed.
\paragraph{Continuous diffusion models.}
For continuous data, the forward process corrupts the initial sample \( \mathbf{x}_0 \) by adding Gaussian noise over \( T \) steps, forming a Markov chain \( \{\mathbf{x}_t\}_{t=0}^T \). Each step introduces noise based on a variance schedule \( \beta_t \):
\begin{equation}
q(\mathbf{x}_t \mid \mathbf{x}_{t-1}) = \mathcal{N}(\mathbf{x}_t; \sqrt{1  -\beta_t} \cdot \mathbf{x}_{t-1}, \beta_t \cdot \mathbf{I}),
\end{equation}
where \( \beta_t \in (0, 1) \). As \( t \) increases, the sample becomes increasingly indistinguishable from pure noise. The noise is gradually accumulated, and we can express the relation between \( \mathbf{x}_t \) and the original sample \( \mathbf{x}_0 \) as:
\begin{equation}
q(\mathbf{x}_t \mid \mathbf{x}_0) = \mathcal{N}(\mathbf{x}_t; \sqrt{\bar{\alpha}_t} \cdot \mathbf{x}_0, (1 - \bar{\alpha}_t) \cdot \mathbf{I}),
\end{equation}
where \( \bar{\alpha}_t = \prod_{s=1}^t (1 - \beta_s) \) represents the cumulative noise added up to time \( t \). This relation shows that as the forward process progresses, the data becomes more corrupted, with \( \bar{\alpha}_t \) diminishing over time.
The reverse process aims to denoise the final noisy sample \( \mathbf{x}_T \) and recover the original sample \( \mathbf{x}_0 \). This process learns a parameterized distribution:
\begin{equation}
p_\theta(\mathbf{x}_{t-1} \mid \mathbf{x}_t),
\end{equation}
which is typically modeled as a Gaussian distribution with learnable mean \( \mu_\theta \) and variance \( \Sigma_\theta \):
\begin{equation}
p_\theta(\mathbf{x}_{t-1} \mid \mathbf{x}_t) = \mathcal{N}(\mathbf{x}_{t-1}; \mu_\theta(\mathbf{x}_t, t), \Sigma_\theta(\mathbf{x}_t, t)).
\end{equation}
During training, \( \mu_\theta \) is optimized to effectively remove noise at each step, while \( \Sigma_\theta \) is often fixed or simplified for stability. By iteratively "denoising" \( \mathbf{x}_t \), the model can recover the original data distribution.

\paragraph{Discrete diffusion models.}
For discrete data, the forward process corrupts the initial sample \( \mathbf{x}_0 \) using Bernoulli noise over \( T \) steps, forming a Markov chain \( \{\mathbf{x}_t\}_{t=0}^T \). Each step flips the values of the data with a probability \( \beta_t \):
\begin{equation}
q(\mathbf{x}_t \mid \mathbf{x}_{t-1}) = \text{Bernoulli}\left( \mathbf{x}_t; (1 - \beta_t) \mathbf{x}_{t-1} + \frac{\beta_t}{2} \right).
\end{equation}
As \( t \) increases, \( \mathbf{x}_t \) becomes increasingly uniform over the discrete space. The distribution of \( \mathbf{x}_t \) is gradually smoothed, with the relationship between \( \mathbf{x}_t \) and \( \mathbf{x}_0 \) given by:
\begin{equation}
q(\mathbf{x}_t \mid \mathbf{x}_0) = \text{Bernoulli}\left( \mathbf{x}_t; \gamma_t \cdot \mathbf{x}_0 + \frac{1 - \gamma_t}{2} \right),
\end{equation}
where \( \gamma_t = \prod_{s=1}^t (1 - \beta_s) \). As \( \gamma_t \to 0 \), \( \mathbf{x}_t \) approaches a uniform distribution over the discrete space.
The reverse process seeks to denoise \( \mathbf{x}_T \) back to \( \mathbf{x}_0 \) by learning the distribution:
\begin{equation}
p_\theta(\mathbf{x}_{t-1} \mid \mathbf{x}_t),
\end{equation}
which is parameterized as a discrete distribution:
\begin{equation}
p_\theta(\mathbf{x}_{t-1} \mid \mathbf{x}_t) = \text{Bernoulli}\left( \mathbf{x}_{t-1}; f_\theta(\mathbf{x}_t, t) \right).
\end{equation}
During training, the model minimizes the cross-entropy loss between the predicted distribution \( p_\theta \) and the true posterior \( q(\mathbf{x}_{t-1} \mid \mathbf{x}_t, \mathbf{x}_0) \), analogous to noise prediction in continuous diffusion models.

\subsection{Multiobjective Reinforcement Learning}
In many real world scenarios, decision-making involves optimizing multiple, often conflicting objectives. Multiobjective reinforcement learning (MORL) \cite{hayes2022practical} extends the standard RL framework by utilizing a vector-valued reward signal
\begin{equation}
    \mathbf{r}_t = [r_t^{(1)}, r_t^{(2)}, \dots, r_t^{(m)}],
\end{equation}
where each component \(r_t^{(i)}\) corresponds to a distinct objective. Improving one objective can degrade performance in another, so specialized algorithms are required to handle these trade-offs effectively.

Formally, an MORL environment is typically represented by a Markov Decision Process (MDP) \(\langle \mathcal{S}, \mathcal{A}, \mathcal{P}, \mathbf{r}, \gamma \rangle\). Here, \(\mathcal{S}\) is the state space, \(\mathcal{A}\) is the action space, \(\mathcal{P}(s' \mid s, a)\) defines the transition probabilities, and \(\gamma \in [0,1)\) is the discount factor. The vector reward function \(\mathbf{r}: \mathcal{S} \times \mathcal{A} \to \mathbb{R}^m\) provides different reward signals for each objective. The goal is to learn a policy \(\pi: \mathcal{S} \to \mathcal{A}\) that balances multiple objectives as encoded in the vector-valued return:
\begin{equation}
\mathbf{G}_t = \sum_{k=0}^\infty \gamma^k \mathbf{r}_{t+k},
\end{equation}
where \(\mathbf{G}_t = [G_t^{(1)}, G_t^{(2)}, \dots, G_t^{(m)}]\) accumulates the rewards for each objective. Because no single solution can optimize all objectives simultaneously, \emph{Pareto optimality} is often used to evaluate policies that cannot be improved in one objective without sacrificing another. Managing these high-dimensional, conflicting objectives remains a significant challenge.
 In MORL \cite{liuefficient,zhu2023scaling}, an agent must handle multiple, often conflicting objectives which can be viewed as different dimensions or “channels” of a complex decision space. By leveraging diffusion-based methods, the proposed LacaDM can generate candidate policies that systematically explore and refine this high-dimensional space, potentially yielding a diverse set of Pareto optimal solutions.

\section{Causal Representation Learning for MORL}

In nonstationary environments, reward dynamics and objective trade-offs may shift over time due to latent factors, making it increasingly challenging for the agent to balance multiple, often conflicting objectives. Effectively modeling the underlying causal structure over these latent factors is essential for achieving robust, generalizable, and adaptive policy behavior.

\textbf{Temporally causal representation modeling.}
We model the causal representation based on latent temporally causal processes \cite{yao2021learning}, which aims to recover latent factors driving temporal dynamics. Formally, let $\mathbf{x}_t$ denote the observed state-action-reward tuple at time $t$. The data generating mechanism can be modeled as: 
\begin{equation}
\mathbf{x}_t = g(\mathbf{z}_t), \quad \mathbf{z}_t \in \mathbb{R}^k
\end{equation}
where $\mathbf{z}_t = (z_{1,t}, \dots, z_{k,t})$ are latent variables evolving via a delayed causal process:
\begin{equation}
z_{i,t} = f_i\left(\{\mathbf{z}_{t-\tau}\}_{\tau=1}^{L}, \epsilon_{i,t} \right), \quad \forall i \in \{1, \dots, k\}
\label{eq:latent_dynamics}
\end{equation}
Here, $f_i$ captures nonlinear causal influences from past latent states, and $\epsilon_{i,t}$ represents exogenous noise due to unobserved environment shifts.
This is corresponded to a temporal causal structure:
$\{\mathbf{z}_{t-\tau}\}_{\tau=1}^{L} \rightarrow \mathbf{z}_t \rightarrow \mathbf{x}_t$.
Under this model, $\mathbf{z}_t$ encodes latent task preferences and environmental dynamics influencing observed behaviors.

\textbf{Policy adaptation via causal inference.}
To enable policy adaptation, we use an encoder-decoder architecture to infer latent variables $\mathbf{z}_t$ and model their dynamics. When a shift $\Delta z_{i,t}$ is detected due to environmental perturbations, the policy input is adjusted proactively:
\begin{equation}
a_t = \pi_\theta\left( \{\mathbf{z}_{t-\tau}\}_{\tau=1}^L + \Delta z_{i,t} \right).
\end{equation}
This allows counterfactual reasoning, i.e., simulating actions under altered causal contexts.

\textbf{Agent and environment effects disentanglement.}
To separate external environment drift from internal policy effects, we model the conditional distribution of the noise term:
\begin{equation}
p(\epsilon_{i,t}) = \frac{\partial s_i(\epsilon_{i,t})}{\partial \epsilon_{i,t}} \cdot \mathcal{N}(s_i(\epsilon_{i,t})),
\end{equation}
where $s_i(\cdot)$ is a normalizing flow transforming $\epsilon_{i,t}$ into a standard distribution. This helps identify whether observed deviations stem from environmental shifts or policy deficiencies.

In summary, causal representation learning (CRL) equips the MORL agent with essential reasoning capabilities that are difficult to achieve with conventional methods. These include the ability to intervene on latent task or environmental factors to synthesize adaptive policies, reason counterfactually about hypothetical situations, and model temporally delayed causal effects in dynamic environments. Together, these capabilities provide a principled and actionable foundation for enhancing generalization, robustness, and adaptability in multi-objective reinforcement learning.

\section{Methodology}
In this section, we first introduce the overview of the proposed LacaDM, and then detail the key components of the LacaDM framework, including forward diffusion for noise estimation, and generation of optimal policies via reverse diffusion.

\subsection{Overview}
\begin{figure*}[!t]
\centering
\includegraphics[width=6.8in]{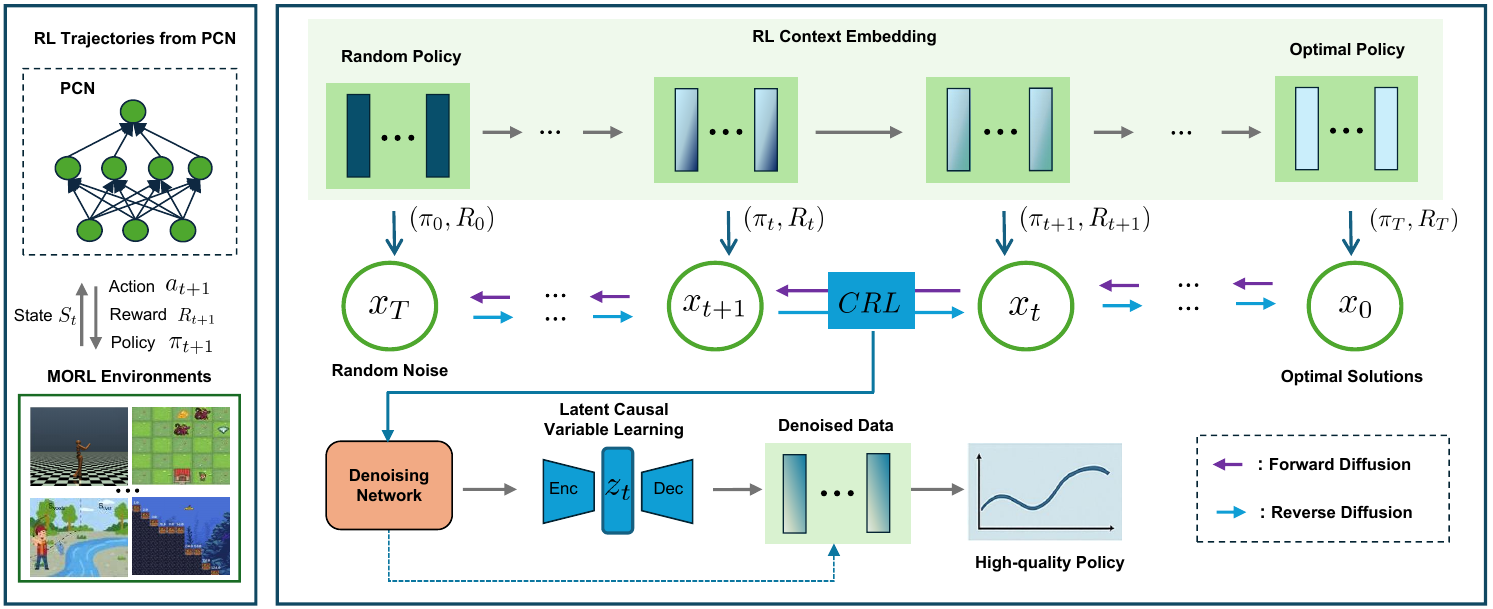}
\caption{Overview of the proposed LacaDM. Inverse RL-guided context embeddings derived from PCN-generated trajectories guide the forward diffusion from optimal solutions to random noise. Latent causal variables \( z_t \), learned via CRL, support reverse denoising to recover high-quality policy solutions.}
\label{fig_1}
\end{figure*}

Figure~\ref{fig_1} illustrates the overall architecture of the proposed LacaDM, which integrates latent causal modeling with a bidirectional diffusion process to enable robust policy learning in MORL environments. The framework comprises a forward diffusion that progressively injects noise into the policy space to promote exploration and diversity, and a reverse diffusion that iteratively removes noise to recover high-quality policies.
To guide the forward diffusion during action generation, we construct an inverse reinforcement learning (IRL) context embedding with historical trajectories from PCN \cite{reymond2022pareto,beliaev2025inverse}.
This embedding captures temporal dependencies within state-action-reward sequences and provides a compact summary of the agent’s past behaviors under varying objective trade-offs.
CRL serves different purposes across the two processes: during forward diffusion, it extracts latent causal variables from observed trajectories, while during reverse diffusion, these variables are used to guide denoising and policy reconstruction.
By modeling causal relationships between extracted latent causal variables, CRL improves LaCaDM's ability to generalize across nonstationary tasks with shifting preferences and dynamics.

\subsection{Forward Diffusion for Noise Estimation}
In the proposed LacaDM, we use a reinforcement learning context embedding to guide the diffusion process, as illustrated in Fig.~\ref{fig_1}. Specifically, the policy at each time step, along with the changes in cumulative rewards, serves as a reinforcement signal that helps LacaDM capture the implicit optimization behavior of different RL algorithms.
We begin by solving \( N \) distinct MORL problems to convergence and recording the resulting policy $\pi_T$ and cumulative reward sequences $R_T$:
\begin{equation}
\pi_T = \left\{ \pi_1, \pi_2, \dots, \pi_t \right\}, \quad 
R_T = \left\{ R_1, R_2, \dots, R_t \right\},
\end{equation}
where \( \pi_t = \left\{ \pi_t(s_1), \pi_t(s_2), \dots, \pi_t(s_n) \right\} \) denotes the policy at time step \( t \), and \( T \) is the total number of time steps used in the diffusion process. These sequences form the conditioning input to our diffusion model.

The forward diffusion process progressively estimates a noise distribution from the policy sequence \( \pi_T \), refining it to match a predefined target distribution. We use a Gaussian noise model in continuous action spaces and a Bernoulli model in discrete environments. At each step \( t \), the noise is injected and updated depending on the environment characteristics, starting from an initial time step \( t = 0 \).
To go beyond pure stochastic degradation and capture environment-specific structure in policy evolution, we introduce a sequence of latent variables \( \{z_t\} \), inferred from the joint trajectories of \( \pi_t \) and \( R_t \). These variables follow a causal generative process:
\begin{equation}
z_t = f(Pa(z_t), \varepsilon_t),
\end{equation}
where \( Pa(z_t) \) are the causal parents from previous steps and \( \varepsilon_t \) represents diffusion noise. These latent variables capture both environment-specific semantics and temporal dependencies. An encoder is trained to infer these latent embeddings based on CRL, allowing LacaDM to model not only noise but also the underlying causal dynamics of policy transitions.

This iterative procedure continues until the noise model converges. The outcome is a pretrained LacaDM model that has internalized both the stochastic degradation and the latent causal structure of MORL processes, enabling effective policy generation in the reverse diffusion phase.



\subsection{Generation of Optimal Policies via Reverse Diffusion}

The reverse diffusion process in LacaDM aims to recover the optimal policy \( \hat{\pi}_0 \) from noise-corrupted inputs by progressively denoising a sequence of latent policy representations \( \hat{\pi}_t \). This process is governed by the following transition model:

\begin{equation}
p(\hat{\pi}_t | \hat{\pi}_{t-1}) = \begin{cases}
\mathcal{N}(\hat{\pi}_t; \mu(\hat{\pi}_{t-1}), \sigma^2), & \text{if continuous} \\
\text{Bernoulli}(\hat{\pi}_t; \pi(\hat{\pi}_{t-1})), & \text{if discrete}
\end{cases}
\end{equation}

Here, \( \mu(\cdot) \), \( \sigma^2 \), and \( \pi(\cdot) \) are predicted by a denoising network, modeling the policy transition dynamics for both continuous and discrete environments. The process proceeds iteratively until a denoised policy \( \hat{\pi}_0 \) is obtained.

\paragraph{Incorporating causal learning.}
To enhance adaptability across diverse MORL environments, we introduce CRL in our LacaDM. At each reverse diffusion step, the denoised policy \( \hat{\pi}_t \) is not only conditioned on its previous state \( \hat{\pi}_{t-1} \), but also influenced by a latent variable \( z_t \) that encodes underlying causal structure, which can be formalized as
\begin{equation}
\hat{\pi}_t = f(\hat{\pi}_{t-1},z_t, \varepsilon_{\hat{\pi}_t}),
\label{equ:cause}
\end{equation}
where $\varepsilon_{\hat{\pi}_t}$ is stochastic noise from the forward diffusion process. 

Based on Eq. (\ref{eq:latent_dynamics}) and Eq. (\ref{equ:cause}), an encoder network is trained to infer \( z_t \) from observed noisy policy trajectories. It maps local segments of policy history to compact latent embeddings, which then guide the reverse diffusion process. The paired decoder \( d_(z_t, \hat{\pi}_{t-1}) \) predicts the denoised policy step, enforcing consistency between causal latent variables and policy evolution. This mechanism enables LacaDM to model not just statistical transitions, but also the structural and temporal causal dynamics behind policy evolution.

\paragraph{CRL-guided policy update.}
Rather than treating reverse diffusion purely as a sampling procedure, we interpret each step as a local policy optimization guided by both denoising accuracy and causal coherence. Specifically, we define a composite loss:

\begin{equation}
\begin{aligned}
\mathcal{L}_{\text{total}}(\hat{\pi}_{t-1}, \hat{\pi}_t, z_t) = 
&\underbrace{\| \mu_\theta(\hat{\pi}_t) - \hat{\pi}_{t-1} \|^2}_{\text{denoising loss}} \\
+ & \beta \cdot \underbrace{\| d(\hat{\pi}_{t-1}, z_t) - \hat{\pi}_t \|^2}_{\text{causal consistency}} 
+  \lambda \cdot \| \hat{\pi}_{t-1} \|_1.
\end{aligned}
\end{equation}
The first term ensures accurate denoising from the noisy trajectory, the second term enforces consistency with the latent causal dynamics via a learned decoder \( d\), and the third introduces \( L_1 \) regularization to encourage sparsity for improved generalization.
We then update the policy using gradient descent:
\begin{equation}
\hat{\pi}_{t-1}^{\text{CRL}} = \hat{\pi}_{t-1} - \alpha \cdot \nabla_{\hat{\pi}_{t-1}} \mathcal{L}_{\text{total}}(\hat{\pi}_{t-1}, \hat{\pi}_t, z_t).
\end{equation}
By embedding this loss-guided update into each reverse diffusion step, LacaDM refines policies not only based on statistical reconstruction but also on latent causal structure. This hybrid learning mechanism improves robustness and adaptability, particularly in scenarios involving domain shifts or temporally evolving objectives.

\section{Experiments}
\subsection{Experiment Setup}

\paragraph{MORL environments.}
To evaluate the performance of LacaDM, we adopt the MOGymnasium framework \cite{felten2023toolkit}, a standardized benchmark suite for MORL. Built on Gymnasium, it supports a wide range of environments with multiple objective functions.
We select eight discrete and eight continuous environments from MOGymnasium. The discrete environments include Deep Sea Treasure, HighwayEnv, ResourceGathering, FourRoom, FruitTree, Breakable Bottles, Fishwood, and MOLunarLander. The continuous environments include MountainCar, WaterReservoir, HopperEnv, MOHalfCheetah, MOAnt, MOSwimmer, MOHumanoid, and MOWalker2D.
This diverse selection enables a comprehensive evaluation of LaCaDM's performance across both discrete and continuous MORL tasks.

\paragraph{Training datasets and Baselines.}
To construct the training dataset for LacaDM, we use the PCN \cite{reymond2022pareto} to solve four MORL environments: Minecart, MOSuperMario, MOReacher, and DeepSeaTreasureMirrored. During agent-environment interaction, we record the state, action, reward, and policy at each time step, resulting in a rich trajectory dataset that captures the temporal dynamics and optimization behavior of multiobjective tasks.
This dataset is used to supervise the forward diffusion process in LaCaDM, enabling the model to capture underlying patterns in policy evolution and multiobjective decision-making.

We compare LacaDM against a diverse set of baselines across both continuous and discrete MORL tasks, using their default hyperparameters. The baselines include two reinforcement learning methods (DQN \cite{mnih2013playing}, PCN \cite{reymond2022pareto}), two evolutionary algorithms (NSGA-III-EHVI \cite{pang2022expensive}, ANSGA-II \cite{liu2022learning}), and three diffusion-based models (EmoDM \cite{yan2024emodm}, MTDiff \cite{he2023diffusion}, DMBP \cite{zhihedmbp}). These baselines cover traditional, evolutionary, and generative approaches, offering a comprehensive benchmark for evaluating LacaDM.

\begin{table*}[htbp]
    \centering
    \caption{Comparison of average HV results in discrete and continuous MORL environments.}
    \label{tab:hv_merged}
    \resizebox{\textwidth}{!}{
    \begin{tabular}{c|c|ccccccccc}
        \toprule
        \textbf{MORL Environment} & \textbf{Env. Type} & \textbf{Deep Qlearning} & \textbf{PCN} & \textbf{ANSGAII} & \textbf{NSGAIIIEHVI} & \textbf{EmoDM} & \textbf{MTDiff} & \textbf{DMBP} & \textbf{LacaDM (Ours)}& \textbf{p-value} \\
        \midrule
        Deep Sea Treasure & Discrete & 2.63e+2 & 3.02e+2 & 2.56e+2 & 3.34e+2 & 2.63e+2 & 3.48e+2 & \textbf{3.52e+2} & 3.50e+2 & 0.000\\
        HighwayEnv        & Discrete & 1.08e+4 & 1.31e+4 & 9.84e+3 & 2.40e+4 & 9.90e+3 & 2.49e+4 & 2.44e+4 & \textbf{2.51e+4} & 0.000 \\
        ResourceGathering & Discrete & 3.83e+0 & 4.74e+0 & 3.94e+0 & 4.70e+0 & 3.95e+0 & 4.80e+0 & 4.81e+0 & \textbf{4.82e+0} & 0.003\\
        FourRoom          & Discrete & 2.13e+1 & 2.25e+1 & 2.24e+1 & 2.56e+1 & 2.18e+1 & 2.55e+1 & \textbf{2.72e+1} & 2.60e+1 & 0.000\\
        FruitTree         & Discrete & 3.34e+4 & 3.31e+4 & 2.98e+4 & 3.45e+4 & 2.84e+4 & 3.57e+4 & 3.61e+4 & \textbf{3.64e+4} & 0.000\\
        BreakableBottles  & Discrete & 2.54e+4 & 2.34e+4 & 2.65e+4 & 2.81e+4 & 2.67e+4 & 2.81e+4 & 2.79e+4 & \textbf{2.82e+4} & 0.100\\
        Fishwood          & Discrete & 3.12e+3 & 3.03e+3 & 2.84e+3 & 3.02e+3 & 2.98e+3 & \textbf{3.17e+3} & 3.05e+3 & 3.15e+3 & 0.000\\
        MOLunarLander     & Discrete & 8.21e+8 & 8.15e+8 & 8.10e+8 & 8.13e+8 & 8.08e+8 & 8.22e+8 & 8.20e+8 & \textbf{8.23e+8} & 0.000\\
        \midrule
        MountainCar       & Continuous & 4.53e+6 & 4.61e+6 & 4.50e+6 & 4.80e+6 & 4.64e+6 & 4.97e+6 & 5.00e+6 & \textbf{5.02e+6} & 0.000\\
        Water Reservoir   & Continuous & 3.12e+5 & 3.24e+5 & 3.06e+5 & 3.08e+5 & 3.07e+5 & 3.18e+5 & 3.42e+5 & \textbf{3.44e+5} & 0.000\\
        HopperEnv         & Continuous & 6.82e+4 & 8.36e+4 & 6.77e+4 & 9.12e+4 & 6.76e+4 & \textbf{9.87e+4} & 9.84e+4 & 9.84e+4 & 0.000\\
        MOHalfcheetah     & Continuous & 6.26e+4 & 6.30e+4 & 6.11e+4 & 6.32e+4 & 6.21e+4 & 6.47e+4 & 6.48e+4 & \textbf{6.50e+4} & 0.000\\
        MOAnt             & Continuous & 1.21e+7 & 1.29e+7 & 1.02e+7 & 1.28e+7 & 1.15e+7 & 1.29e+7 & 1.29e+7 & \textbf{1.31e+7} & 0.000\\
        MOSwimmer         & Continuous & 1.24e+4 & 1.26e+4 & 9.98e+3 & 1.31e+4 & 1.02e+4 & 1.48e+4 & 1.50e+4 & \textbf{1.53e+4} & 0.000\\
        MOHumanoid        & Continuous & 2.00e+5 & 1.92e+5 & 1.65e+5 & 2.02e+5 & 1.74e+5 & \textbf{2.26e+5} & 2.24e+5 & 2.21e+5 & 0.000\\
        MOWalker2D        & Continuous & 5.44e+4 & 5.42e+4 & 5.01e+4 & 5.51e+4 & 5.05e+4 & 5.48e+4 & 5.55e+4 & \textbf{5.67e+4} & 0.001\\
        \bottomrule
    \end{tabular}
    }
    
\end{table*}

\subsection{Results and Performances}
\paragraph{Hypervolume performances.}
Table \ref{tab:hv_merged} shows the Hypervolume (HV) values of LacaDM and baseline models in discrete and continuous MORL environments, respectively. The HV metric measures the accuracy and diversity of the solution set, by calculating as the volume of the hypercube between the Pareto Front (PF) and a reference point. We use the default reference points provided by the MOGymnasium framework. A higher HV value indicates better overall algorithm performance.

In Table \ref{tab:hv_merged}, LacaDM achieved the highest average HV values in five out of eight scenarios in the discrete environments, and in six out of eight scenarios in the continuous environments.
The results are based on the average of 10 independent runs, with statistical tests confirming that LacaDM outperformed the baseline models at a significance level of $p < 0.05$ after 10 runs.
In the discrete environments, LacaDM particularly excelled in complex scenarios such as Breakable Bottles and MOLunarLander, where it significantly outperformed the baseline models.
These results underscore LacaDM’s ability to handle environments with intricate reward structures and sparse objective distributions. Similarly, in the continuous environments, LacaDM achieved optimal performance in high-dimensional tasks like MOHalfcheetah, MOSwimmer and MOWalker2D, showcasing its capability to manage the challenges of continuous action spaces and dynamics. The superior HV values achieved by LacaDM across various environments validate its potential for solving complex MORLs. Notably, when compared to MTDiff and DMBP, two reinforcement learning diffusion-based models, LacaDM exhibited clear advantages in both continuous and discrete MORL tasks. This further underscores the effectiveness of our causal  representation learning mechanism in enhancing the diffusion process and improving optimization performance across diverse MORL scenarios.

\paragraph{Sparsity performances.}
To further evaluate the performance of our LacaDM, we assess sparsity, which measures how evenly the solutions are distributed across the objective space. Lower sparsity values indicate better performance. The p-values in Table \ref{tab:s_all_environments} further confirm the statistical significance of these results.
As shown in Table \ref{tab:s_all_environments}, LacaDM achieved the lowest average sparsity values in five out of eight discrete environments and six out of eight continuous environments, based on the average of 10 independent runs. These results highlight the effectiveness of LacaDM’s diffusion process in generating evenly distributed solutions. Notably, LacaDM performed exceptionally well on high-dimensional continuous tasks, such as MO-Halfcheetah and MO-Walker2D, and complex discrete tasks like MO-Lunar-Lander, where capturing causal dynamics is critical for balancing solution distribution. Overall, LacaDM consistently maintains low sparsity values, demonstrating that causal representation learning, combined with the diffusion process, plays a key role in producing diverse and evenly distributed solution sets, a critical requirement for effective multi-objective optimization.

\begin{table*}[htbp]
    \centering
        \caption{Comparison of average Sparsity across discrete and continuous MORL environments.}
    \resizebox{\textwidth}{!}{
    \begin{tabular}{c|c|ccccccccc}
        \toprule
        \textbf{MORL environments} & \textbf{Env. Type} & \textbf{Deep Qlearning} & \textbf{PCN} & \textbf{ANSGAII} & \textbf{NSGAIIIEHVI} & \textbf{EmoDM} & \textbf{MTDiff} & \textbf{DMBP} & \textbf{LacaDM (Ours)}& \textbf{p-value} \\
        \midrule
        Deep Sea Treasure & Discrete & 1.53e+1 & 1.45e+1 & 2.01e+1 & 1.35e+1 & 1.97e+1 & 1.34e+1 & 1.28e+1 & \textbf{1.24e+1} & 0.000\\
        HighwayEnv        & Discrete & 1.83e+1 & 1.67e+1 & 2.15e+1 & 1.94e+1 & 2.20e+1 & \textbf{1.41e+1} & 1.68e+1 & 1.65e+1 & 0.000\\
        ResourceGathering & Discrete & 8.09e+2 & 6.42e+2 & 1.25e+1 & 8.02e+2 & 1.03e+1 & 6.42e+2 & 6.35e+2 & \textbf{6.33e+2} & 0.000\\
        FourRoom          & Discrete & 3.37e+2 & 3.42e+2 & 5.21e+2 & 3.30e+2 & 5.17e+2 & 3.18e+2 & \textbf{3.11e+2} & 3.15e+2 & 0.000\\
        FruitTree         & Discrete & 2.40e+1 & 2.41e+1 & 3.17e+2 & 2.35e+2 & 3.26e+2 & 2.08e+2 & \textbf{1.98e+2} & 2.02e+2 & 0.000\\
        BreakableBottles  & Discrete & 3.54e+1 & 3.65e+1 & 4.02e+1 & 3.55e+1 & 4.00e+1 & 3.29e+1 & 3.27e+1 & \textbf{3.26e+1} & 0.000\\
        Fishwood          & Discrete & 1.58e+0 & 1.66e+0 & 2.21e+0 & 1.54e+0 & 2.18e+0 & 1.48e+0 & 1.52e+0 & \textbf{1.43e+0} & 0.000\\
        MOLunarLander     & Discrete & 1.25e+1 & 1.26e+1 & 1.45e+1 & 1.22e+1 & 1.48e+1 & 1.19e+1 & 1.18e+1 & \textbf{1.18e+1} & 0.001\\
        \midrule
        MountainCar       & Continuous & 7.57e+1 & 6.12e+1 & 9.04e+1 & 7.51e+1 & 8.97e+1 & 6.21e+1 & 6.18e+1 & \textbf{6.02e+1} & 0.000\\
        Water Reservoir   & Continuous & 2.14e+0 & 1.95e+0 & 3.45e+0 & 2.01e+0 & 3.33e+0 & \textbf{1.88e+0} & 1.94e+0 & 1.92e+0 & 0.000\\
        HopperEnv         & Continuous & 1.46e+0 & 1.24e+0 & 3.14e+0 & 1.20e+0 & 3.24e+0 & 1.11e+0 & 1.08e+0 & \textbf{1.03e+0} & 0.000\\
        MOHalfcheetah     & Continuous & 9.36e+0 & 9.42e+0 & 1.26e+1 & 9.23e+0 & 1.22e+1 & 9.21e+0 & 9.18e+0 & \textbf{9.15e+0} & 0.000\\
        MOAnt             & Continuous & 2.66e+2 & 2.48e+2 & 3.54e+2 & 2.40e+2 & 3.28e+2 & 2.17e+2 & 2.20e+2 & \textbf{2.15e+2} & 0.000\\
        MOSwimmer         & Continuous & 9.16e+0 & 9.23e+0 & 1.54e+1 & 9.02e+0 & 1.02e+1 & \textbf{8.45e+0} & 8.57e+0 & 8.77e+0 & 0.000\\
        MOHumanoid        & Continuous & 5.72e+1 & 5.87e+1 & 6.25e+1 & 5.69e+1 & 6.11e+1 & 5.67e+1 & 5.87e+1 & \textbf{5.55e+1} & 0.002\\
        MOWalker2D        & Continuous & 2.31e+2 & 2.28e+2 & 2.94e+2 & 2.22e+2 & 2.88e+2 & 2.04e+2 & 2.11e+2 & \textbf{2.01e+2} & 0.000\\
        \bottomrule
    \end{tabular}
    }
    \label{tab:s_all_environments}
\end{table*}

\begin{figure*}[h!]
    \centering
    \includegraphics[width=0.23\textwidth]{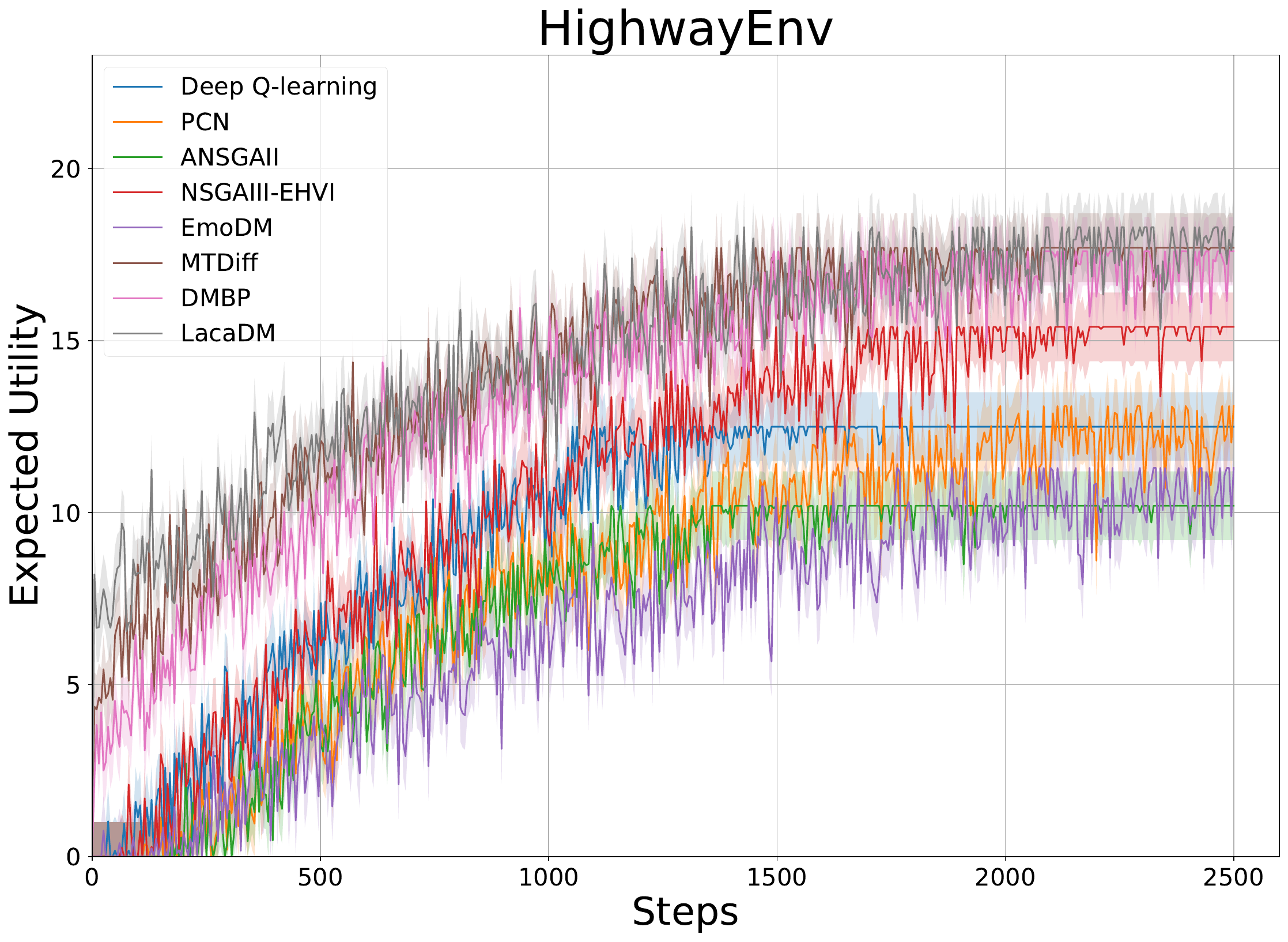}
    \includegraphics[width=0.23\textwidth]{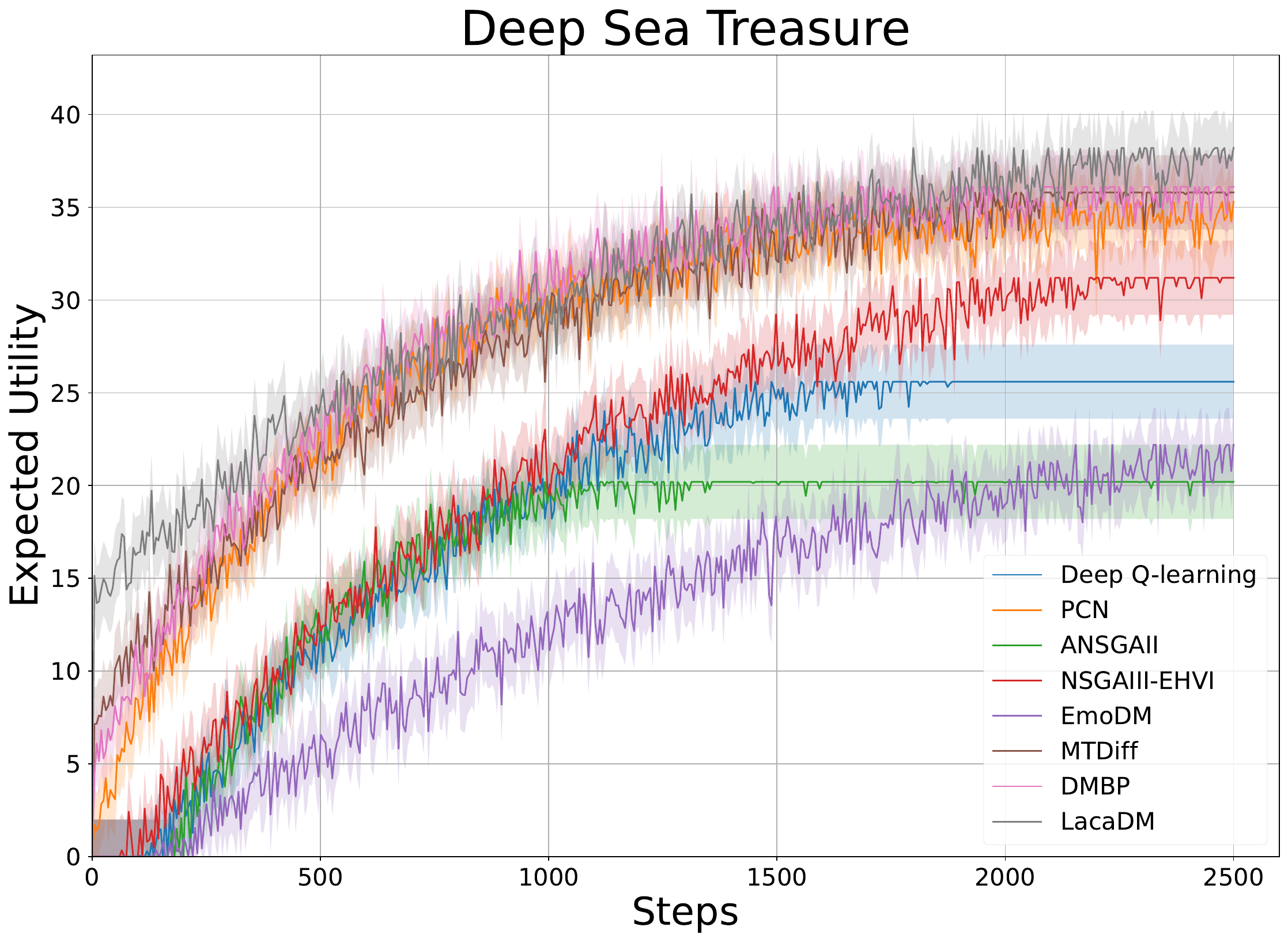}
    \includegraphics[width=0.23\textwidth]{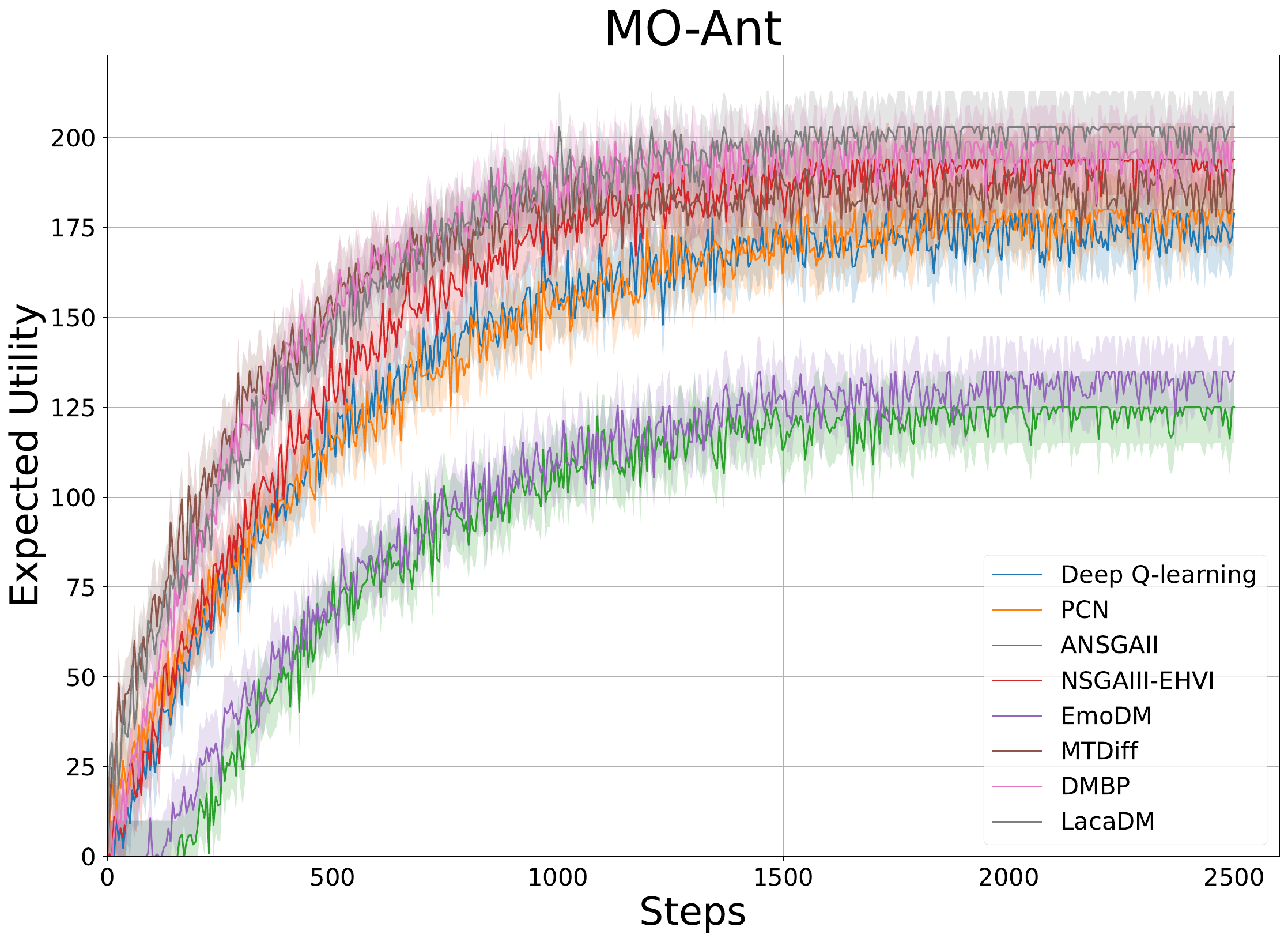}
    \includegraphics[width=0.23\textwidth]{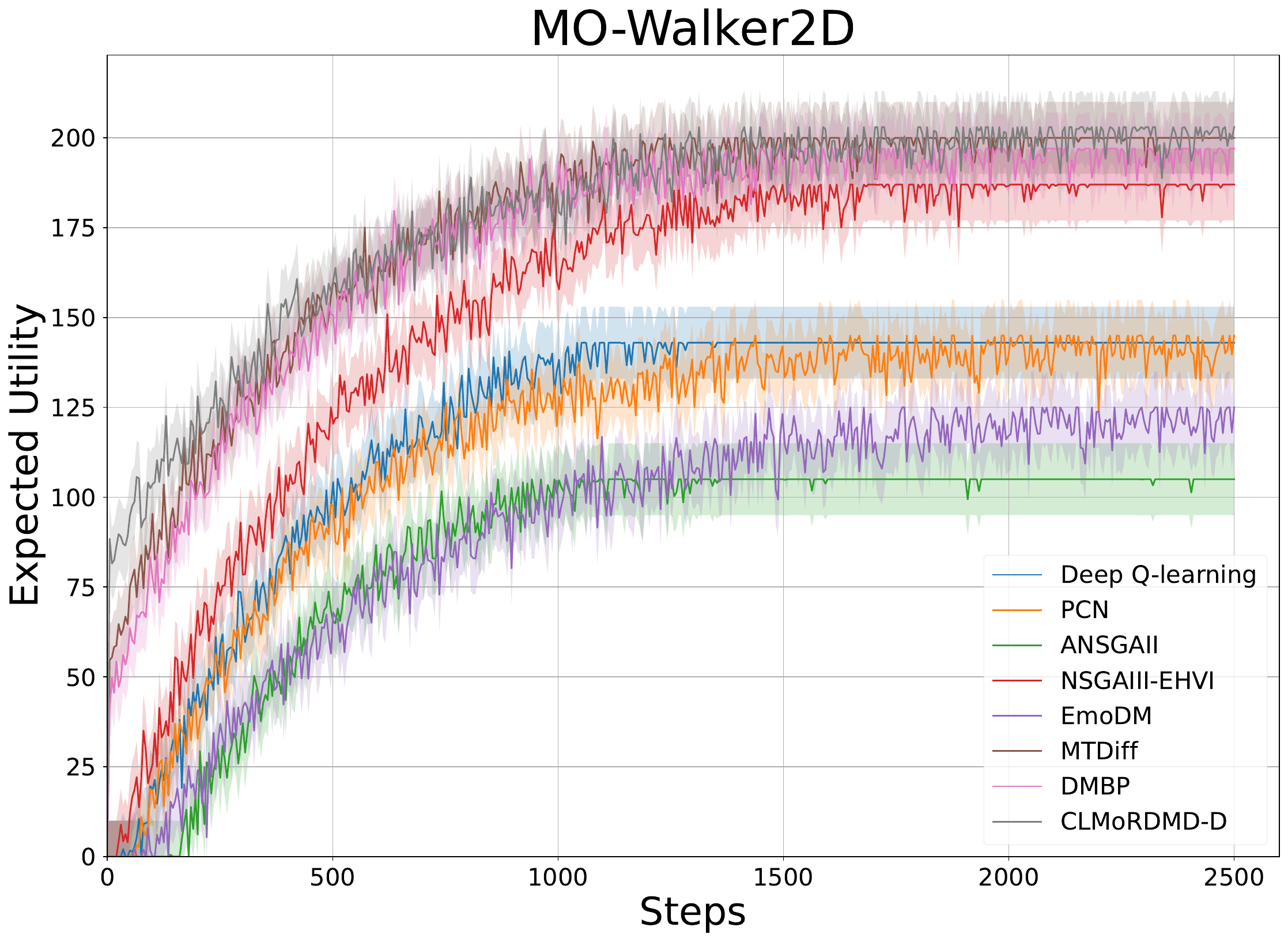}
    \caption{Expected utility of baseline models and LacaDM across four MORL problems as the number of solving steps increases.}
    \label{fig:expected}
\end{figure*}

\paragraph{Expected utility maximization performances.}
 We also use Expected Utility Maximization (EUM) as an evaluation metric. EUM reflects the decision-maker's preferences by calculating a weighted average of utility values across all possible outcomes, where larger values indicate better performance. For each environment, we use the default implementation from the MO-Gymnasium framework to compute EUM values.
 Figure \ref{fig:expected} illustrates the trends of expected utility over training steps for HighwayEnv and Deep Sea Treasure (discrete environments) and MO-Ant and MO-Walker2D (continuous environments). The results consistently show that LacaDM outperforms baseline models across all tasks.
In discrete environments, LacaDM demonstrates rapid convergence, achieving higher EUM values than baselines within the first 1000 steps and maintaining its lead throughout training. In continuous environments, LacaDM achieves superior performance by consistently reaching higher EUM values earlier in training. Its performance stabilizes near the optimal solution after 1500 steps, highlighting its ability to efficiently explore and converge in challenging continuous control tasks.
 
These results reflect LacaDM’s advantages in both discrete and continuous settings, where faster convergence and better stability stem from its key mechanisms. Specifically, causal representation  learning captures causal relationships between objectives, guiding the reverse diffusion process for more efficient exploration and convergence. Meanwhile, the diffusion process achieves a balance between solution diversity and optimization through the stepwise introduction and removal of noise, enabling LacaDM to generate high-quality strategies more effectively than other methods.



    
\subsection{Effect of Latent Casual Learning }
In this section, we conduct experiments on a variant of LacaDM, LacaDM-CRL (which removes the CRL component), to analyze and validate the effectiveness of causal representation Learning (CRL). To evaluate its performance, we selected four environments: Fishwood and HighwayEnv from the discrete set, 
and HopperEnv and MountainCar from the continuous set. All other experimental settings remain consistent with the original LacaDM configuration.
To isolate the impact of the CRL component in LacaDM, we compared the results of LacaDM with those of LacaDM-CRL. Table \ref{tab:lcl_ablation} presents the comparative results in these four environments, demonstrating the critical role of CRL in the policy generation process. Without CRL, LacaDM loses the ability to capture the potential relationships between the new environment and the predicted noise during strategy generation. Consequently, the reverse diffusion process fails to converge effectively, preventing the model from reaching the optimal strategy.
\begin{table}{}{}
\centering
\caption{Hypervolume (HV) results of LacaDM with and without the CRL component across four MORL environments.}
\label{tab:lcl_ablation}
   \resizebox{0.45\textwidth}{!}{
\begin{tabular}{lcc}
\toprule
\textbf{Environment} & \textbf{LacaDM-CRL} & \textbf{LacaDM} \\
\midrule
Fishwood     & 2.42e+3 & \textbf{3.15e+3 }\\
HighwayEnv   & 9.87e+3 & \textbf{2.51e+4} \\
MountainCar  & 4.24e+6 & \textbf{5.02e+6} \\
HopperEnv    & 6.38e+4 & \textbf{9.84e+4} \\
\bottomrule
\end{tabular}
}
\end{table}

\begin{figure}{}{}  
    \centering
    \includegraphics[width=0.18\textwidth]{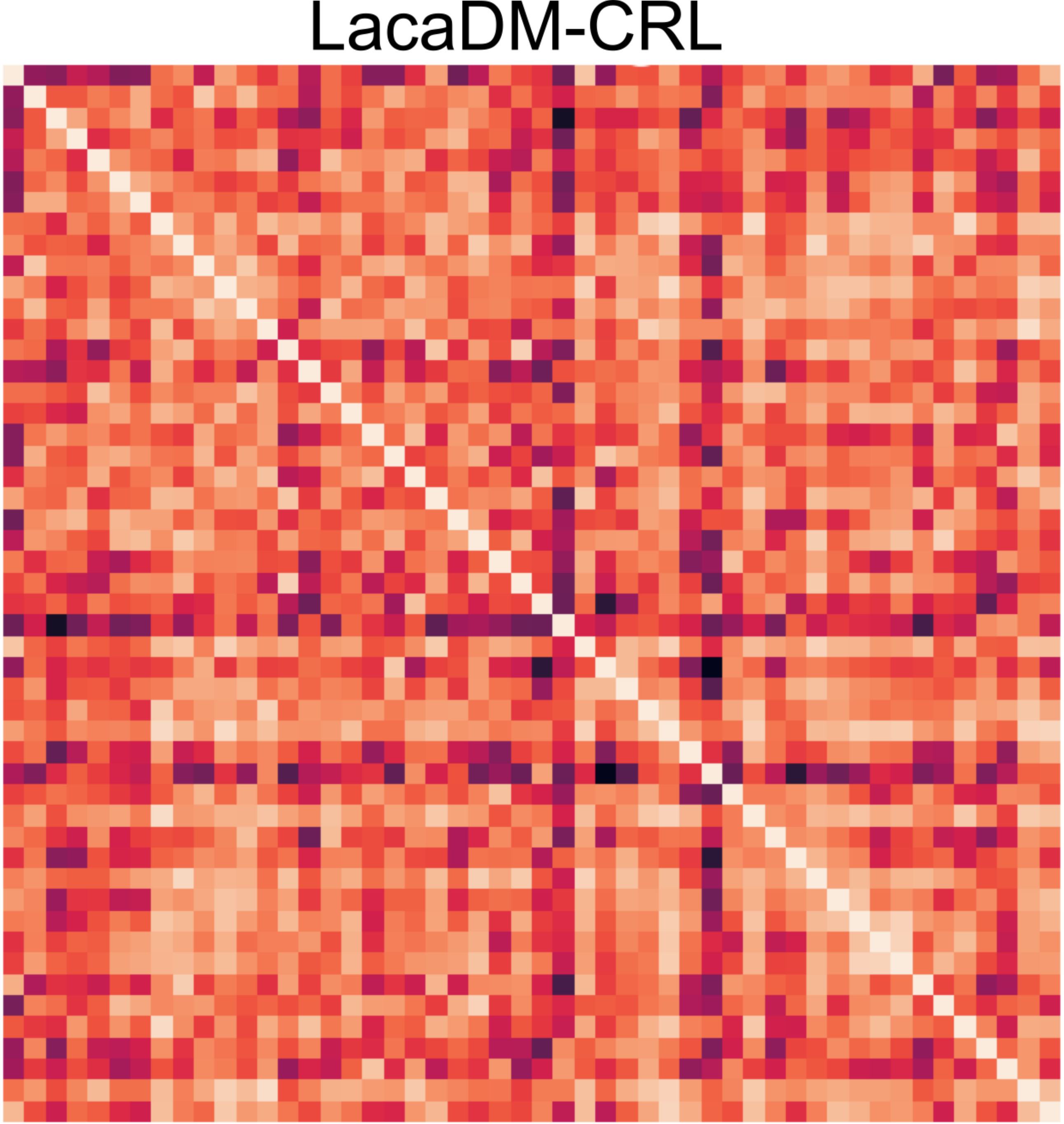}
     \hspace{0.02\textwidth} 
    \includegraphics[width=0.18\textwidth]{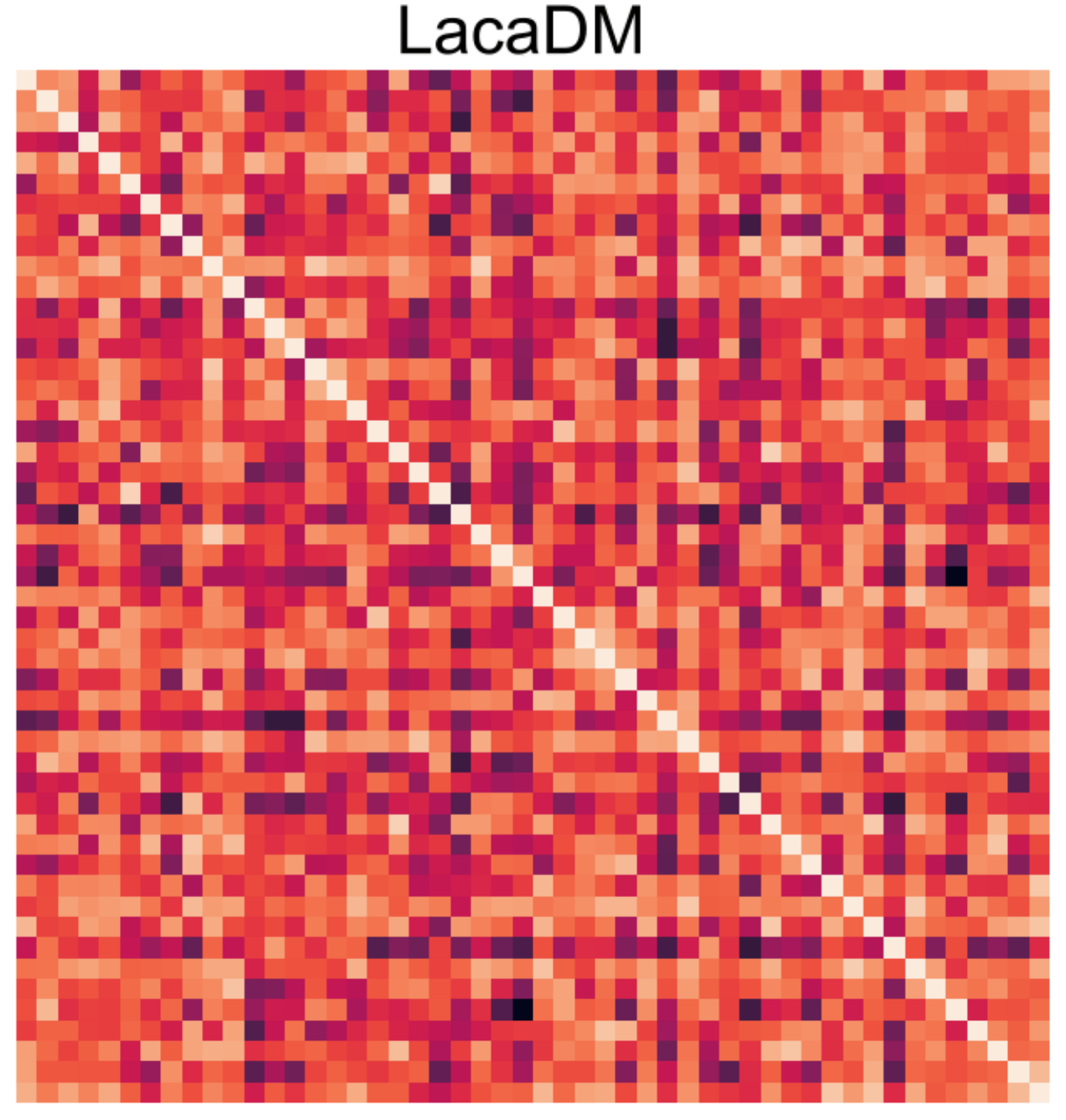}
  \caption{Cosine similarity heatmaps comparing noise inference and training between LacaDM-CRL and LacaDM at the midpoint of the diffusion process.}
  \label{fig:cos}
  \end{figure}

In addition, to investigate whether CRL is the key factor contributing to this phenomenon, we generated cosine similarity heatmaps of noise inference and training at \(\frac{T}{2}\), with and without the CRL component. In our experiment, \(T = 1500\), the training environment for the sampled data was MO-SuperMario, and the reasoning environment was MO-Walker2D. The resulting heatmaps are shown in Figure \ref{fig:cos}.  
As illustrated in Figure \ref{fig:cos}, the overall color of the heatmap is noticeably darker when the CRL component is included compared to when it is not. This indicates that our proposed model with the CRL component is better able to generalize from the training environment to an unseen reasoning environment. These results highlight the critical role of CRL in improving the model's ability to adapt and transfer knowledge across different environments.  
\section{Conclusion}
In this paper, we proposed LacaDM, a novel diffusion model enhanced with CRL to tackle complex MORL problems. LacaDM combines the strengths of diffusion models for effective exploration and convergence with the power of CRL to capture dynamic relationships between objectives and environments.
This integration enables LacaDM to produce diverse and high-quality solutions across discrete and continuous MORL environments.
Through extensive experiments on MO-Gymnasium environments, we demonstrated that LacaDM consistently outperforms baseline methods
, including reinforcement learning algorithms, evolutionary algorithms, and existing diffusion-based approaches, 
in both discrete and continuous tasks. These results highlight the effectiveness of CRL in improving the diffusion model's ability to generalize across environments and optimize strategies efficiently. 
In the future, we plan to explore more advanced causal inference techniques for the diffusion model to further enhance the scalability and generalization capabilities of LacaDM. Additionally, we also aim to explore the integration of other advanced techniques such as multi-agent collaboration and transfer learning to further push the boundaries of LacaDM in solving even more complex MORL tasks.




\section*{Ethical Statement}

There are no ethical issues.


\bibliographystyle{named}
\bibliography{ijcai26}

\end{document}